\documentclass[10pt,twocolumn,letterpaper]{article}

\usepackage{cvpr}
\usepackage{times}
\usepackage{epsfig}
\usepackage{graphicx}
\usepackage{amsmath}
\usepackage{amssymb}
\usepackage{subcaption}
\usepackage{graphicx}
\usepackage{multirow}
\usepackage{xcolor}
\usepackage{colortbl}

\usepackage[breaklinks=true,bookmarks=false]{hyperref}

\cvprfinalcopy % *** Uncomment this line for the final submission

\def\cvprPaperID{****} % *** Enter the CVPR Paper ID here

\begin{document}

%%%%%%%%% TITLE
\title{ Enhanced Object Detection: A Study on Vast Vocabulary Object Detection Track for V3Det Challenge 2024 }

\author{Peixi Wu \\
University of Science and Technology of China\\
{\tt\small wupeixi@mail.ustc.edu.cn}
\and
Bosong Chai\thanks{Bosong Chai is the corresponding author. Bosong Chai and Peixi Wu contributed equally to this work.}\\
Zhejiang University\\
{\tt\small chaibosong@mail.zju.edu.cn}
\\
\and
Xuan Nie\\
Northwestern Polytechnical University\\
{\tt\small xnie@nwpu.edu.cn}
\\
\and
Longquan Yan\\
Northwest University\\
{\tt\small 18829512640@163.com}
\\
\and
Zeyu Wang\\
Zhejiang University\\
{\tt\small wangzeyu2020@zju.edu.cn}
\\
\and
Qifan Zhou\\
Northwestern Polytechnical University\\
{\tt\small george13@mail.nwpu.edu.cn}
\and
Boning Wang\\
Zhejiang University\\
{\tt\small 1007658022@qq.com}
\\
\and
Yansong Peng \\
University of Science and Technology of China\\
{\tt\small pengyansong@mail.ustc.edu.cn}
\\
\and
Hebei Li \\
University of Science and Technology of China\\
{\tt\small lihebei@mail.ustc.edu.cn}}

\maketitle

\begin{abstract}
  In this technical report, we present our findings from the research conducted on the Vast Vocabulary Visual Detection (V3Det) dataset for Supervised Vast Vocabulary Visual Detection task. How to deal with complex categories and detection boxes has become a difficulty in this track. The original supervised detector is not suitable for this task. We have designed a series of improvements, including adjustments to the network structure, changes to the loss function, and design of training strategies. Our model has shown improvement over the baseline and achieved excellent rankings on the Leaderboard for both the Vast Vocabulary Object Detection (Supervised) track and the Open Vocabulary Object Detection (OVD) track of the V3Det Challenge 2024.

\end{abstract}

%%%%%%%%% BODY TEXT
\section{Introduction}
\begin{figure}[ht]
    \centering
    \begin{minipage}[b]{0.32\linewidth}
        \centering
        \includegraphics[width=\linewidth]{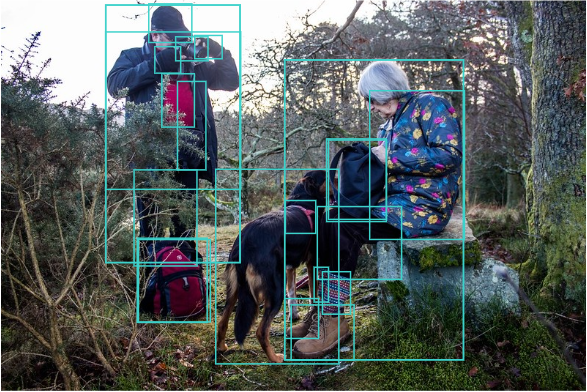}
    \end{minipage}
    \hfill
    \begin{minipage}[b]{0.32\linewidth}
        \centering
        \includegraphics[width=\linewidth]{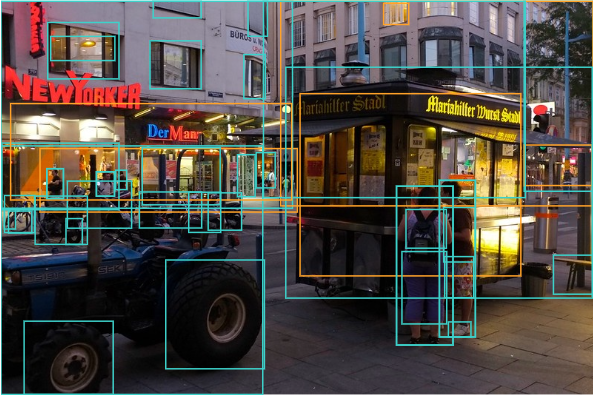}
    \end{minipage}
    \hfill
    \begin{minipage}[b]{0.32\linewidth}
        \centering
        \includegraphics[width=\linewidth]{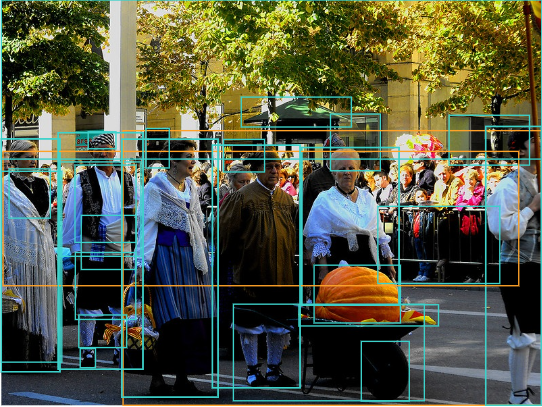}
    \end{minipage}
    \caption{V3Det is a high-quality, precisely annotated object detection dataset with a broad vocabulary, encompassing 13,204 categories. The figure shows annotated image samples from V3Det, featuring more complex and detailed annotations.}
    \label{fig1}
\end{figure}

The V3Det dataset~\cite{wang2023v3det} is a large-scale, richly annotated dataset featuring detection bounding box annotations for over 13,000 object classes on real images. It includes a hierarchical category structure with detailed class affiliations forming a comprehensive relationship tree. As shown in Fig~\ref{fig1}, with 245,000 annotated images and expert-generated descriptions, V3Det is an invaluable resource for advanced object detection research in computer vision.

% \begin{figure}[ht]
%     \centering
%     \begin{subfigure}[b]{0.32\linewidth} 
%         \centering
%         \includegraphics[width=\linewidth]{fig1.jpg}
%         % \caption{}
%     \end{subfigure}
%     \hfill
%     \begin{subfigure}[b]{0.32\linewidth} 
%         \centering
%         \includegraphics[width=\linewidth]{fig3.jpg}
%         % \caption{}
%     \end{subfigure}
%     \hfill
%     \begin{subfigure}[b]{0.32\linewidth} 
%         \centering
%         \includegraphics[width=\linewidth]{fig2.jpg}
%         % \caption{}
%     \end{subfigure}
%     \caption{V3Det is a high-quality, precisely annotated object detection dataset with a broad vocabulary, encompassing 13,204 categories. The figure shows annotated image samples from V3Det, featuring more complex and detailed annotations.}
%     \label{fig1}
% \end{figure}

This workshop has two tracks. The first track (Supervised), called Vast Vocabulary Object Detection, aims to evaluate supervised learning models for object detection across all 13,204 classes in the V3Det dataset. Detecting any object has been a long-term goal in the field of computer vision. Due to the countless diverse objects in the real world, an ideal visual detection system should be capable of detecting a large number of categories and be applicable to open vocabulary categories. 

Currently widely used object detection datasets such as COCO~\cite{lin2014microsoft}, Objects365~\cite{shao2019objects365}, and OpenImages v4~\cite{kuznetsova2020open}, despite providing a large number of images and categories, still have a limited vocabulary. The limited vocabulary of these datasets constrains the training potential of class-generalized detectors, as an ideal detector should be able to recognize new categories beyond those in the training set. Even large vocabulary object detection datasets like LVIS~\cite{gupta2019lvis} cannot fully represent the complexity of the real world in terms of the number and diversity of categories. V3Det provides the research community with a large vocabulary object detection dataset, which can accelerate the exploration of more general visual detection systems. The baseline cascade structure is very suitable for handling the hierarchical category structure of the V3Det dataset. We treat the supervised track \uppercase\expandafter{\romannumeral1} as a traditional object detection task with complex labels, using common detection improvement strategies. By improving the  Feature Pyramid Network (FPN) structure, we hope the network can effectively learn deeper semantic information. Additionally, we balance category labels by adjusting the loss function. 

The second track (OVD) of the V3Det challenge involves developing object detectors capable of accurately identifying objects from 6,709 base classes and 6,495 novel classes. For base classes, full annotations are provided, while for novel classes, only class names, descriptions, and a few exemplar images are given. The task is to design detectors that can utilize this limited information to detect novel classes effectively during inference, ensuring accurate detection across both base and novel categories. This track requires detectors to possess strong generalization and semantic understanding capabilities to identify new categories without direct annotation information. It can rely on current vision-text models, such as CLIP~\cite{radford2021learning}, to extract visual and semantic features from images and text, and establish connections between them. 

The baseline EVA model~\cite{fang2023eva}, combined with CLIP~\cite{radford2021learning}, demonstrates powerful semantic feature extraction capabilities. Due to time constraints and limited computational resources, we rely solely on supervised training for Track \uppercase\expandafter{\romannumeral2}, yet still achieve good detection results even for novel categories. This to some extent indicates that V3Det dataset covers a vast array of annotations from real-world scenarios, with rich semantic information learned by excellent detectors, thus exhibiting good generalization performance.

%------------------------------------------------------------------------
\section{Related Work}

\subsection{Object Detection}
Object detection~\cite{girshick2015fast,girshick2014rich,cai2018cascade} is one of the most traditional tasks in computer vision, with various applications across different industries such as autonomous driving~\cite{wu2023transformation,wang2023tirdet,mao20233d}, robotics~\cite{coates2010multi}, remote sensing~\cite{chai2024enhanced}. It takes images as input, localizes, and classifies objects within a given vocabulary. Each detected object is represented by a bounding box with a class label. 

Classical CNN-based object detectors can be divided into two main categories: two-stage and one-stage detectors. Two-stage detectors~\cite{girshick2015fast,girshick2014rich,cai2018cascade,zhou2021probabilistic} first generate object proposals and then refine them in a second stage, offering higher precision but at the cost of increased complexity. One-stage detectors, such as YOLO~\cite{redmonyolo9000,9823408,wang2022yolov7,wang2024yolov10} and SSD\cite{liu2016ssd}, directly classify and regress predefined anchor boxes or search for geometric cues like points~\cite{tian1904fcos}, centers~\cite{duan2019centernet}, and corners~\cite{law2018cornernet}, providing faster but potentially less accurate results. Transformer-based detectors~\cite{sun2021rethinking,he2022destr,carion2020end,zhu2020deformable} use the self-attention mechanism to capture global contextual information in images, eliminating the need for additional components like anchor boxes and Non-Maximum Suppression (NMS). The end-to-end architecture is simpler, making the training and inference process more straightforward. 

Currently, novel detectors based on diffusion are emerging~\cite{chen2023diffusiondet,chen2023diffusion}. At the same time, object detection is being combined with large language models (LLM) to achieve open-vocabulary detection~\cite{wu2023cora,cheng2024yolo,wang2023detecting} and the detection of everything. This approach allows object detection to go beyond just the design of detector architectures, providing models with better adaptability to handle complex scenes and various types of objects.

\subsection{Data Augmentation}
Data augmentation is a commonly used technique in machine learning and deep learning, aimed at transforming and expanding training data to increase its diversity and richness. In addition to common data augmentation methods such as flipping, jittering, and scaling, effective data augmentation techniques for object detection can be broadly categorized into Cutting-based~\cite{zhong2020random,devries2017improved} and Mixing-based~\cite{zhang2017mixup,yun2019cutmix,kim2020puzzle} methods. There is also the widely used Mosaic method proposed by YOLOv4~\cite{bochkovskiy2020yolov4}.

%-------------------------------------------------------------------------

\section{Our Method}
In this section, we elaborate on the technical details of our method. We made two improvements based on the baseline: (a) adjustments to the model architecture, (b) improvements to the loss function and training strategy. We will introduce each component in the following subsections.

\subsection{Baseline Framework}
In this challenge, the organizers built two baselines based on MMDetection~\cite{chen2019mmdetection}\footnote{\href{https://github.com/open-mmlab/mmdetection}{https://github.com/open-mmlab/mmdetection}} and Detectron2\footnote{\href{https://github.com/facebookresearch/detectron2}{https://github.com/facebookresearch/detectron2}}. 
The baseline EVA\footnote{\href{https://github.com/V3Det/Detectron2-V3Det}{Detectron2-V3Det-EVA}}, based on Detectron2, utilizes a Cascade RCNN with a backbone structure of ViTDet~\cite{li2022exploring}. The pretraining task of EVA involves Masked Image Modeling (MIM), aimed at reconstructing masked image-text aligned visual features generated by the CLIP~\cite{radford2021learning}. This network demonstrates robust generalization performance and stands as the state-of-the-art (SOTA) for many vision tasks. Based on the MMDetection baseline\footnote{\href{https://github.com/V3Det/mmdetection-V3Det/tree/main/configs/v3det}{MMDetection-V3Det}}, the best-performing model is also based on Cascade R-CNN~\cite{cai2018cascade}, with a Swin-Transformer~\cite{liu2021swin} as its backbone. The cascade structure is highly suitable for multi-class detection tasks by progressively refining bounding boxes and classification results. Each stage of the cascade head uses two shared fully connected layers, which helps capture high-level semantic features of the targets at different stages. The IoU thresholds set for each stage ensure that the detection boxes become more precise at each level.

\subsection{Model Architecture Adjustment}

\noindent
\textbf{Backbone.}
The baseline adopts Swin Transformer~\cite{liu2021swin} as the backbone network for feature extraction, commonly using versions such as Swin-S, Swin-B, and Swin-L\footnote{{\href{https://github.com/microsoft/Swin-Transformer}{https://github.com/microsoft/Swin-Transformer}}}. Different versions affect the parameter count, computational cost, and accuracy. Therefore, we have made multiple attempts with different backbones. The baseline pretrained model provided by the organizers uses ImageNet-1K pretrained weights to initialize the backbone. We also attempted to use ImageNet-22K pretrained weights to initialize the Swin-B backbone. We also attempted to use pretrained models with a resolution of 384×384\footnote{{\href{https://github.com/SwinTransformer/storage/releases/download/v1.0.0/swin_base_patch4_window12_384_22k.pth}{swin\_base\_patch4\_window12\_384\_22k.pth}}}. In addition to using Swin Transformer as the backbone, we also experimented with the basic Vision Transformer models, specifically using ViT-B and ViT-L.

\begin{figure}
  \begin{center}
  \includegraphics[width=1.6in]{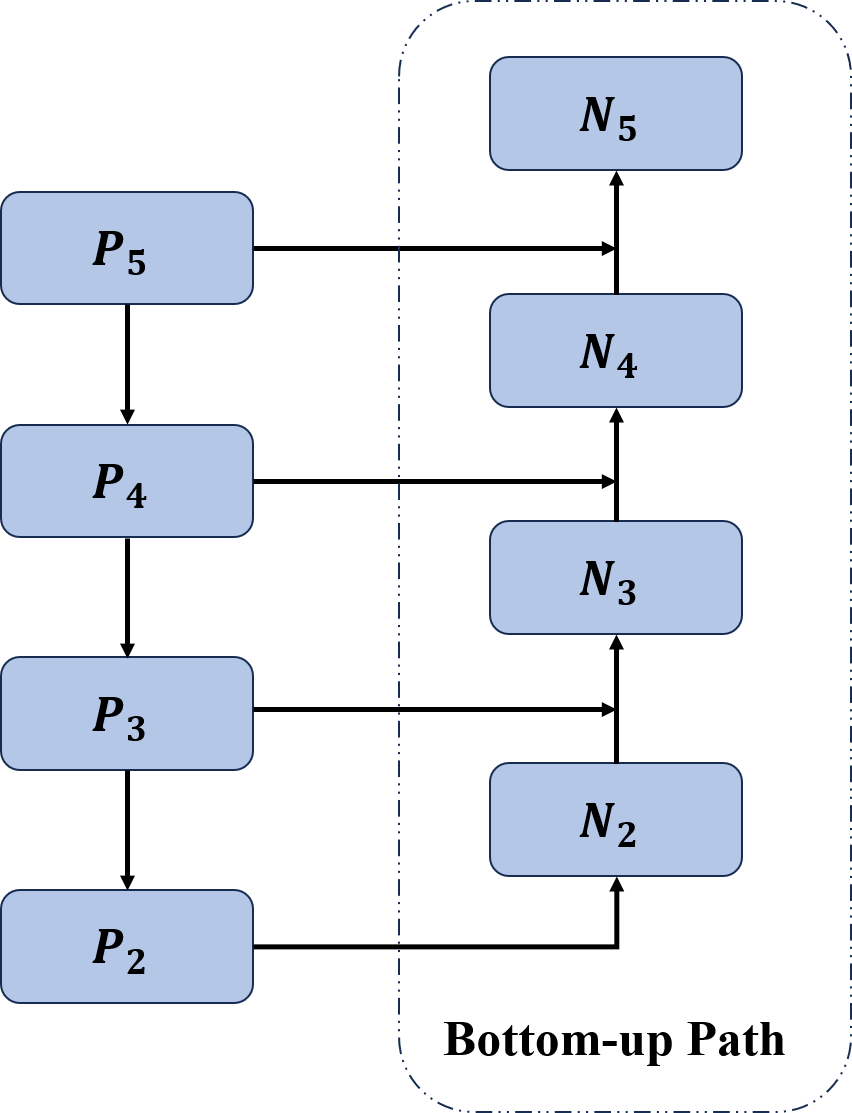}\\
  \caption{ Illustration of PA-FPN structure, FPN with bottom-up path structure from $N_2$ to $N_5$.}
  \label{fig.4}
  \end{center}
\end{figure}

\noindent
\textbf{Path Aggregation Feature Pyramid Network (PA-FPN).}
Although the FPN structure already integrates shallow feature information, the path from shallow features to the top network layers is too long, resulting in low utilization efficiency of shallow features. To effectively capture image semantic information, inspired by PA-Net~\cite{liu2018path}, we add a bottom-up structure into the baseline Cascade R-CNN. This shortens the transmission path from shallow features to the top layers, enhancing the transmission of shallow features within the network, and allowing more shallow features to be effectively utilized. As shown in Fig~\ref{fig.4}, where the feature map $N_2$ has the same dimensions as $P_2$. {$N_3$, $N_4$, $N_5$} are obtained through downsampling and fusion. For a high-resolution feature map $N_i$ and a low-resolution feature map $P_{i+1}$, a new feature map $N_{i+1}$ is generated.

\subsection{Other Improvements}

\noindent
\textbf{Data Augmentation.}
In order to enhance the size and quality of training dataset, we employ data augmentation including flipping, jittering, and scaling, on original input images. We tried the data augmentation strategies built into MMDetection-transforms such as Mixup, Cutout, Corrupt, and PhotoMetricDistortion. It is important to note that more data augmentation is not always better, especially in object detection tasks. Excessive data augmentation can lead to shifts or distortions in the original target positions, making it difficult for the model to learn accurate target boundaries. It has been shown~\cite{szegedy2016rethinking} that the two-stage algorithm can be used for data augmentation without random geometric transformations in the training phase.

\begin{table*}
\small
	\centering
	\caption{ The detection results of different models on the V3Det Supervised track \uppercase\expandafter{\romannumeral1}. We show the results on the validation set, with gray indicating the baseline provided by the organizers in MMDetection. The last row represents the baseline provided in Detectron2, which uses EVA pretrained with CLIP. $pretrain$ indicates whether the models are pretrained on the ImageNet 1K or the ImageNet 22K dataset, and $resolution$ indicates whether an input resolution of 224×224 or 384×384 was used. All models are based on Cascade R-CNN. }
	\begin{tabular}{c c c c c c c c c c c c c}
		\hline
		Backbone & pretrain & resolution & $AP_{all}$ & $AP_{50}$ & $AP_{75}$ & $AP_{s}$ & $AP_{m}$ & $AP_{l}$ & Recall\_{s} & Recall\_{m} & Recall\_{l} & Recall\_{all}\\[2.5pt]
		\hline
         \rowcolor{gray!20} Swin-B  & 1K  & 224  & 43.4 & 50.2 & 45.8 & 12.9 & 22.9 & 49.0 & 23.3 & 37.0 & 70.1 & 64.3  \\[1.5pt]
         Swin-B  & 22K  & 224  & 43.2 & 50.4 & 46.0 & 13.7 & 23.7 & 49.3 & 30.6 & 44.8 & 75.3 & 69.5  \\[1.5pt]
         Swin-B  & 22K  & 384  & 43.7 & 50.6 & 46.3 & 13.7 & 23.9 & 49.5 & 30.8 & 45.1 & 75.5 & 69.8  \\[1.5pt]
         Swin-L    & 22K  & 224  & 37.1 & 43.4 & 38.3 & 8.6 & 16.3 & 35.2 & 24.4 & 38.2 & 64.5 & 58.5  \\[1.5pt]
        \hline
         ViT-B  & 22K  & 384  & 40.2 & 46.6 & 43.3 & 10.2 & 19.5 & 40.2 & 30.8 & 40.7 & 68.1 & 69.8  \\[1.5pt]
         ViT-L  & 22K  & 224  & 30.1 & 35.9 & 32.7 & 9.8 & 17.1 & 35.5 & 23.4 & 37.4 & 70.3 & 64.2 \\[1.5pt]
         \hline
         \multicolumn{3}{c}{Cascade R-CNN EVA-CLIP }  & 51.1 & 55.9 & 53.2 & 24.4 & 34.6 & 56.2 & 44.3 & 56.2 & 78.6 & 75.3 \\[1.5pt]
        \hline
	\end{tabular}
	\label{tab_1}
\end{table*}

\noindent
\textbf{Loss Function.}
In this section, we introduce the DIoU Loss function for addressing coordinate point interrelationship issues using the $L_1$ loss function in baseline Cascade R-CNN networks. Inspired by Zhaohui Zheng et al.~\cite{zheng2020distance}, DIoU Loss considers two key issues: (a) Minimizing the normalized distance between the prediction frame and the target frame to achieve faster convergence. (b) How to make the regression more accurate and faster when there is overlap or even inclusion with the target box. The DIoU Loss function yields values in the range [-1,1], and is defined as follows:

\begin{equation}
\small
R_{DIoU}=\frac{\rho^2(b,b^{gt})}{c^2},    
\end{equation}

\begin{equation}
\small
	L_{DIoU}=1-IoU+R_{DIoU}, 
\end{equation}

$\rho(\cdot)$ represents the Euclidean distance. The penalty term $R_{DIoU}$ is defined as the squared Euclidean distance between the central points of $b$ and $b_{gt}$, normalized by the square of the diagonal length $c$ of the smallest enclosing box covering the two boxes. This formulation ensures that the DIoU loss directly minimizes the distance between the two central points.

Inspired by Li et al.~\cite{li2020generalized}, to reduce the economic imbalance of the sample measure in the detection process and the inaccurate detection results caused by the blurred bounding box, we properly introduces the Generalized Focal Loss (GFL) function into the Region Proposal Network (RPN) to balance the proportion of positive and negative samples in the loss function,  The GFL function is typically shown in equation (\ref{6}).

\begin{equation}\label{6}
\small
\begin{aligned}
G F L\left(p_{y_l}, p_{y_r}\right) & =-\left|y-\left(y_l p_{y_l}+y_r p_{y_r}\right)\right|^\beta \\
& \times\left(\left(y_r-y\right) \log \left(p_{y_l}\right)+\left(y-y_l\right) \log \left(p_{y_r}\right)\right).
\end{aligned}
\end{equation}

$y$ represents the true IoU, while $y_l$ and $y_r$ are the lower and upper bounds of the predicted and true IoU of the bounding boxes. $\beta$ is an adjustable hyper-parameter controlling the slope of the loss function ($\beta \geq 0$). $p_{yl}$ and $p_{yr}$ are the probability values predicted by the model, satisfying $p_{y_l}+p_{y_r}=1$. The final prediction $\hat{y}$ is a linear combination of $y_l$ and $y_r$, enabling classification values to transition from discrete to continuous. The balancing factor in the formula minimizes deviations between predicted and true IoU, while the classification loss function computes errors to enhance the model's understanding of object position and size. GFL employs a focal mechanism, dynamically adjusting weights to balance proportions and facilitate learning differences between positive and negative samples.

\begin{table}[h]
\small
	\centering
	\caption{ Detection results of different methods for Supervised track \uppercase\expandafter{\romannumeral1} in the validation set }
	\begin{tabular}{c c c }
        \hline
         Method & $AP_{all}$ &  Recall\_{all} \\[1.5pt]
        \hline
        Baseline  & 43.4  & 64.3   \\[1.5pt]
        
         +PA-FPN  & 42.2  & 62.6    \\[1.5pt]
         +DIoU  & 44.7  & 69.3    \\[1.5pt]
         +GFL  & 43.7  & 68.4    \\[1.5pt]
         
         \hline
	\end{tabular}
	\label{tab2}
\end{table}

\begin{table}[h]
\small
	\centering
	\caption{ Detection results on the test set for OVD Track \uppercase\expandafter{\romannumeral2}, where $n$ represents novel categories and $b$ represents base categories. }
	\begin{tabular}{c c c c c c c c }
        \hline
          $bAP_{50}$ &  $nAP_{50}$ & $bAP_{75}$ & $nAP_{75}$ & $bAP$ & $nAP$ & $AP$ \\[1.5pt]
        \hline
        56.2  & 28.7  & 53.2 & 2.2 & 50.4 & 10.3 & 20.2   \\[1.5pt]
         
         \hline
	\end{tabular}
	\label{tab3}
\end{table}

\noindent
\textbf{Training Techniques.}
During training, we find that the $json$ format files of more than 30 images in the original dataset do not match the corresponding images. We perform data cleaning and remove such erroneous data. We use Synchronized Batch Normalization to solve the multiple GPU cross-card synchronization problem. For the learning rate setting, we borrowed the training strategy of YOLOv3~\cite{redmon2018yolov3}, and in the first 3000 iterations, we use warm-up to gradually increase the learning rate from 0 to the preset base learning rate, and subsequent iterations with the cosine strategy, which is conducive to the stability of the training process. We use Apex-based hybrid precision training to accelerate the training with as little loss of precision as possible. We also enable auto-scale learning rate, which means that when using different numbers of GPUs and different batch sizes, GPU resources can be effectively utilized and the model can converge quickly.

\section{Experiments}
In this section, we present the implementation details and give main experimental results and analysis.

\subsection{Implementation Details}
Following the challenge guidelines, 183,354 images are used as the training set, and 29,821 images are used as the validation set. We train exclusively  on the V3Det dataset and do not use any extra data. We train the full models on the training set and evaluate them on the validation set for algorithm validation and hyper-parameter tuning. Finally, we retrain and save the models on the complete training data using the selected hyper-parameters. We implement our model using PyTorch 2.1.0 and conduct our experiments on a system with 4 × H100 GPUs, using a batch size of 48. We use Adam with decoupled weight decay (AdamW)~\cite{loshchilov2017decoupled} with a learning rate of 0.001. We adopt the COCO Detection Evaluation~\cite{lin2014microsoft} to measure the performance. The COCO Detection Evaluation includes multiple-scale objects ($AP_{S}$, $AP_{L}$), where $AP_{S}$ represents small object $AP$, with an area $<$ 32², and $AP_{L}$ represents large object $AP$, with an area $>$ 96². For the Supervised Track \uppercase\expandafter{\romannumeral1}, $AP$ and Recall are used as evaluation metrics for the test set. For the OVD Track \uppercase\expandafter{\romannumeral2}, $AP$ and Recall are calculated separately for the base categories and novel categories.

\subsection{Results and Analysis}
As shown in Table~\ref{tab_1}, we are trying various approaches to the model backbone. When using ImageNet 22k pretraining, there is not much change in the $AP$ value of the model, but the Recall has significantly improved. The Recall\_{all} has increased from 64.3\% to 69.5\%, indicating that the model misses fewer targets. Better pretraining initialization of the backbone is particularly important for object detection tasks. Using a larger model like Swin-L as the backbone introduces additional parameters and computational complexity, resulting in longer inference times. However, despite these drawbacks, the detection performance of the model decreased. 

\begin{figure}
  \begin{center}
  \includegraphics[width=3.3in]{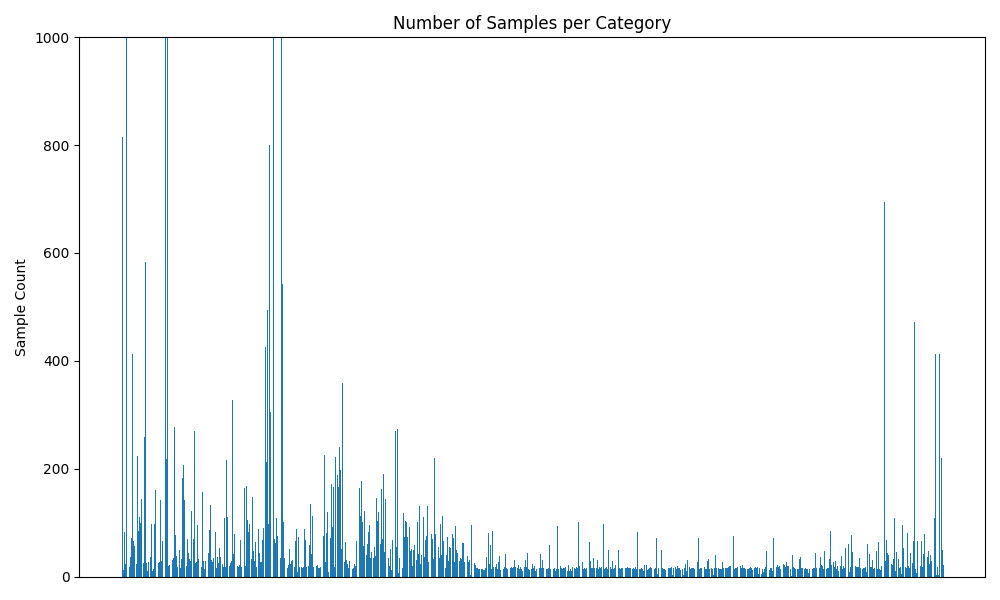}
  \caption{ The horizontal axis represents different classes, and the vertical axis represents the number of samples corresponding to each class, with values above 1000 not displayed.}
  \label{fig.5}
  \end{center}
\end{figure}

As shown in Table~\ref{tab2}, we introduced a series of improvements, including optimizing the loss function of the original detector and modifying its FPN structure. Surprisingly, after incorporating the PA-FPN structure, the model's detection performance, as measured by $AP$, did not improve but instead decreased by nearly 2\%. The PA-FPN structure has been proven effective in many tasks and widely applied in various detection and segmentation tasks. We speculate that this unexpected result may be due to the influence of noise or irrelevant information on the lower-level features, leading to a decrease in the quality of the fused features. The bottom-up structure may cause premature or excessive fusion of features between different levels, resulting in information loss or confusion. The introduction of the bottom-up structure may increase the complexity of the network, making training more challenging and requiring more adjustments and optimizations. Due to time constraints, we did not conduct detailed experiments, and further validation will be carried out gradually.

Certainly, modifying the RPN classification loss function to the GFL function and changing the bounding box regression loss to the GIoU loss function have proven effective. As shown in Fig~\ref{fig.5}, the V3Det dataset, due to its numerous categories, results in poor learning performance for minority classes during training. GFL introduces adjustable parameters to weight the loss functions for different classes, allowing the model to focus more on challenging samples.GFL introduces adjustment parameters to weight the loss functions of different categories, making the model pay more attention to samples that are difficult to classify. 

Regrettably, despite conducting numerous experiments and adjustments, and achieving some improvements over the baseline, our results still could not surpass the reproduced EVA model provided by the organizers based on Detectron2. The EVA model employed the MIM training method, optimizing CLIP and demonstrating powerful performance and superior results. The outstanding performance of the EVA model indicates that merely modifying and designing the model structure is no longer sufficient to achieve significant breakthroughs in the current era of large models. The key to the success of the EVA model lies in its innovative training methods and the effective utilization of pretrained models, which provides a direction for our future research and improvements. 

As shown in Table~\ref{tab2}, for OVD Track \uppercase\expandafter{\romannumeral2}, we adhered to the traditional supervised object detection transfer learning approach and did not incorporate textual information. According to the competition requirements, we used the Cascade R-CNN model based on MMDetection with Swin-B as the backbone from Track \uppercase\expandafter{\romannumeral1}, retrained on the V3Det train set of base classes, and directly inferred on the test dataset. We were pleasantly surprised to find that this approach also yielded good results. Compared to the baseline, our $AP$ for novel classes improved from 11\% to 20\%, with $AP_{50}$ reaching 29\%. This might be because the V3Det dataset already contains rich semantic information, giving the model a certain degree of generalization ability.

\section{Conclusion}
In conclusion, this report has presented our study on V3Det Challenge for Vast Vocabulary Object Detection track 2024. In the Supervised Track \uppercase\expandafter{\romannumeral1}, we made various attempts at traditional object detection tasks using different models. For the V3Det dataset, which contains rich semantic information across multiple categories, we observed some improvement in detection results. However, although the performance did not fully meet our expectations, our adjustments could not surpass the results we obtained by reproducing EVA. This indicates that simply modifying and designing model structures is no longer sufficient in the era of LLM. Our final submission achieved good results on the leaderboard for both Track \uppercase\expandafter{\romannumeral1} and Track \uppercase\expandafter{\romannumeral2}.

% %-------------------------------------------------------------------------

% %-------------------------------------------------------------------------

% %-------------------------------------------------------------------------

% %-------------------------------------------------------------------------

% \clearpage

{\small
\bibliographystyle{ieee}
\bibliography{egbib}
}

\end{document}